\documentclass[preprint]{elsarticle}
\usepackage{graphicx}
\usepackage{physics}
\usepackage{amsmath}
\usepackage{tikz}
\usepackage{mathdots}
\usepackage{yhmath}
\usepackage{cancel}
\usepackage{color}
\usepackage{siunitx}
\usepackage{array}
\usepackage{multirow}
\usepackage{amssymb}
\usepackage{gensymb}
\usepackage{tabularx}
\usepackage{extarrows}
\usepackage{booktabs}
\usetikzlibrary{fadings}
\usetikzlibrary{patterns}
\usetikzlibrary{shadows.blur}
\usetikzlibrary{shapes}
\usepackage{diagbox}
\usepackage{bbding}
\usepackage{amsthm}
\usepackage{float}
\usepackage{bm}

\theoremstyle{definition}

\journal{Pattern Recognition}

\begin{document}

\begin{frontmatter}

\title{Deep Generative Domain Adaptation with Temporal Attention for Cross-User Activity Recognition}

\author[inst1]{Xiaozhou Ye\corref{cor1}\fnref{fn1}}
\ead{xye685@aucklanduni.ac.nz}

\author[inst1]{Kevin I-Kai Wang\corref{cor2}}
\ead{kevin.wang@auckland.ac.nz}
\cortext[cor2]{Corresponding author}

\affiliation[inst1]{organization={Department of Electrical, Computer, and Software Engineering, The University of Auckland}, city={Auckland}, country={New Zealand}}

\begin{abstract}
In Human Activity Recognition (HAR), a predominant assumption is that the data utilized for training and evaluation purposes are drawn from the same distribution. It is also assumed that all data samples are independent and identically distributed ($\displaystyle i.i.d.$). Contrarily, practical implementations often challenge this notion, manifesting data distribution discrepancies, especially in scenarios such as cross-user HAR. Domain adaptation is the promising approach to address these challenges inherent in cross-user HAR tasks. However, a clear gap in domain adaptation techniques is the neglect of the temporal relation embedded within time series data during the phase of aligning data distributions. Addressing this oversight, our research presents the Deep Generative Domain Adaptation with Temporal Attention (DGDATA) method. This novel method uniquely recognises and integrates temporal relations during the domain adaptation process. By synergizing the capabilities of generative models with the Temporal Relation Attention mechanism, our method improves the classification performance in cross-user HAR. A comprehensive evaluation has been conducted on three public sensor-based HAR datasets targeting different scenarios and applications to demonstrate the efficacy of the proposed DGDATA method. 
\end{abstract}


\end{frontmatter}

\section{Introduction}

Human Activity Recognition (HAR) is an integral and essential part of human-computer interaction, ubiquitous computing, and the Internet of Things. Its primary aim is to accurately classify human activities through sensor data and contextual information \cite{dang2020sensor}. Despite considerable research efforts, achieving an effective HAR model remains challenging. A major issue in current HAR methods, particularly those handling time series sensor data, is the assumption that training and testing data come from identical distributions, a premise often invalidated by real-world data diversity and out-of-distribution ($\displaystyle o.o.d. $) instances \cite{chen2021deep}. This mismatch, stemming from factors like sensor heterogeneity \cite{xing2018enabling}, evolving data patterns  \cite{lu2018learning}, different sensor layouts \cite{rokni2018autonomous}, and individual behavioral variances \cite{xiaozhou2023temporaloptimal}, significantly hinders the generalizability of HAR models. Our research is specifically geared towards tackling the $\displaystyle o.o.d. $ challenge in sensor-based HAR arising from individual behavioral differences, known as the cross-user HAR problem, which commonly occurs in healthcare applications.

Transfer learning and domain adaptation \cite{farahani2021brief} offer promising solutions for this problem. These approaches aim to bridge the gap between data distributions of different domains. Techniques like Maximum Mean Discrepancy \cite{yan2017mind} and Covariance Alignment \cite{sun2016return} are employed to align domain features, while more advanced methods such as Transferable Semantic Augmentation \cite{li2021transferable} and Simultaneous Semantic Alignment Network \cite{li2020simultaneous} focus on deeper, non-linear alignments. However, these methods typically assume that data within each domain are independent and identically distributed ($\displaystyle i.i.d.$), an assumption that does not hold for time series data due to their inherent temporal relations. This gap in existing domain adaptation strategies highlights the need for approaches that can effectively utilize these temporal relationships, especially in the context of cross-user HAR.

\begin{figure}[h!]
\centering
\includegraphics[width=0.6\columnwidth]{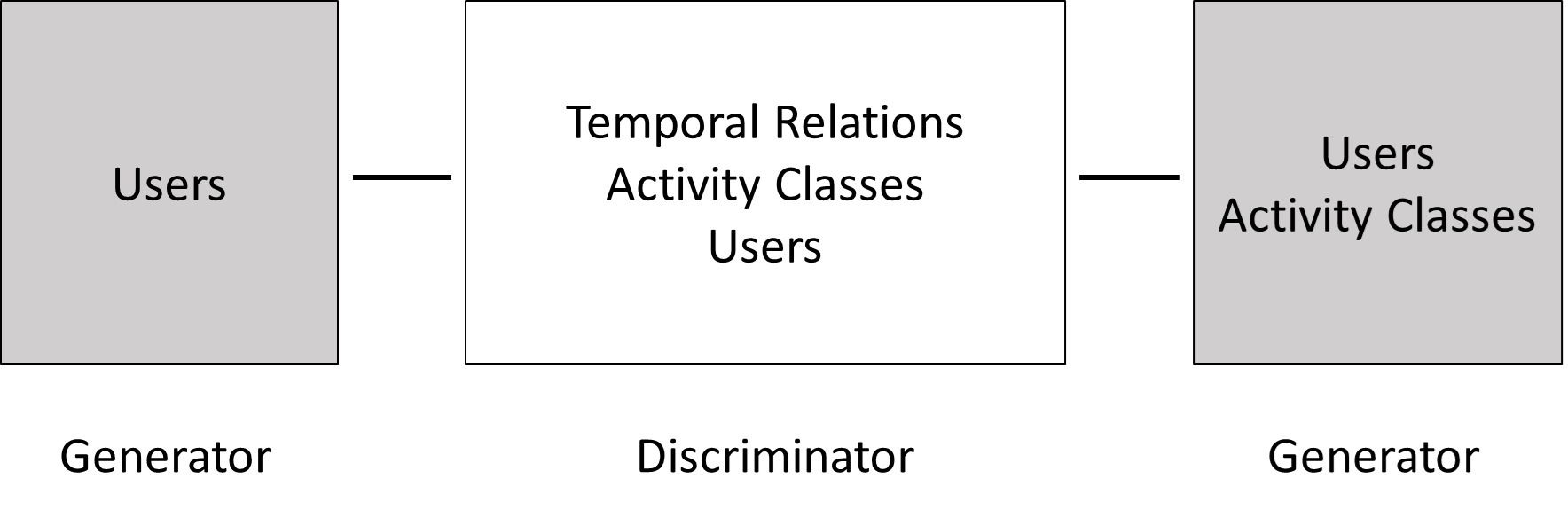}
\caption{The design principle overview of DGDATA.\label{framework_DGDATA}}
\end{figure}

To address this, we introduce a novel method called Deep Generative Domain Adaptation with Temporal Attention (DGDATA), meticulously crafted for time series domain adaptation, in the context of cross-user activity recognition. This approach considers the temporal relation that is hidden in time series data and utilizes it for better data distribution alignment between source and target users. In detail, a Temporal Relation Attention mechanism is proposed where an autoregressive model learns attention to weigh the significance of time steps in a sequence, prioritizing those with the most relevant temporal relations for data distribution alignment. Moreover, the generative model of Conditional Variational Autoencoder (CVAE) is applied as a strategy for better cross-user model generalization considering its capability of capturing underlying data distributions and complex data structures \cite{ramchandran2023learning}. Furthermore, DGDATA utilizes adversarial learning framework, akin to Generative Adversarial Networks (GANs), to extract user-independent temporal relations and lean user-invariant classifier as shown in Figure~\ref{framework_DGDATA}. It features a discriminator (white part) and two generators (grey parts). First, the discriminator and one of the generators cooperate to confuse user and activity class data distributions to extract common temporal relations across users. Then, the discriminator and the other generator work together to aid in creating a user-invariant activity classifier with the help of the extracted common temporal relations. This approach helps in learning an activity classifier that is consistent across different users with the utilization of temporal relation knowledge.
        
In summary, the contributions of our work are specifically:

\begin{enumerate}
\item A new deep domain adaptation framework, DGDATA, is designed to target cross-user HAR with time series data. The framework design absorbs the idea of GAN adversarial learning. Moreover, it combines adversarial learning, temporal attention, and generative modelling in a unique way to utilize temporal relation knowledge embedded in time series data.

\item A novel generative model of CVAE architecture is proposed. It introduced unique constraint conditions and network structure that benefit the data distribution alignment between source and target domains in time series data for better domain adaptation, which enhances the generalized capability of the CVAE model.

\item A novel Temporal Relation Attention mechanism is proposed to capture the importance of time steps in a sequence, considering the common and shared temporal relation across users for better data distribution alignment in cross-user HAR task.
\end{enumerate}

The paper is organized as follows: It begins with a review of related work in cross-user HAR and foundational concepts in transfer learning and domain adaptation. Then, it delves into the specifics of the DGDATA method, its design principles, and its focus on temporal relations. Following this, we discuss our experimental setup and compare DGDATA with existing methods, highlighting its efficacy in utilizing temporal relations for time series data. The paper concludes with a reflection on our findings and suggestions for future research directions in this area.

\section{Related work}

\subsection{Cross-user human activity recognition}

HAR has become a pivotal element in the realm of ubiquitous computing, offering vital insights into human behaviors through the analysis of sensor data and contextual information. This field utilizes diverse sensor types to recognize activities in various contexts. Based on sensor modalities, HAR can be categorized into five main groups: those involving smartphones and wearables, ambient sensors, device-free sensors, vision-based sensors, and other types, as highlighted in several studies \cite{chen2021deep}\cite{lentzas2020non}\cite{zhang2021privacy}\cite{roche2021multimodal}\cite{pan2020hierarchical}. Our study particularly concentrates on HAR via wearable sensors.

In the domain of machine learning, sensor-based HAR is treated as a time series classification issue \cite{yadav2021review}. Various classification models, such as Bayesian network \cite{liu2018learning}, Support Vector Machines \cite{shuvo2020hybrid}, and Hidden Markov Models (HMM) \cite{liu2021motion}, have been proposed to address this challenge. Moreover, deep learning techniques have demonstrated remarkable performance in many tasks. In HAR, deep learning-based approaches \cite{liu2023distributional} \cite{sezavar2024dcapsnet} can autonomously learn and extract features from large datasets. However, these approaches often assume that the training and testing data are from the same distribution, maintaining the model's generalizability under the independent and identically distributed ($\displaystyle i.i.d.$) framework \cite{chen2021deep}. This assumption, however, is frequently invalid in real-world scenarios where training and testing datasets are often out-of-distribution ($\displaystyle o.o.d.$) \cite{lu2022out}. Our study delves into the sensor-based HAR issue within the $\displaystyle o.o.d.$ context.

The sensor-based HAR $\displaystyle o.o.d. $ problem manifests in various forms. First, data may come from different sensors, resulting in varied formats and distributions due to differences in types, platforms, manufacturers, and modalities \cite{xing2018enabling}. Second, data patterns might evolve over time, known as concept drift \cite{lu2018learning}, like changes in walking patterns due to health alterations. Third, significant behavioral differences exist among individuals \cite{saeedi2018personalized}, such as variations in walking speeds. Lastly, the physical placement of sensors, whether on different body parts \cite{rokni2018autonomous} or in smart homes \cite{sukhija2019supervised}, influences data distributions. Our research primarily addresses the challenge of behavioral differences among individuals in sensor-based HAR $\displaystyle o.o.d. $ scenarios.

\subsection{Transfer learning and domain adaptation}

Transfer learning allows models to be trained with one or more source domains and thereafter be deployed in related target domains where labelling might be sparse or non-existent. This a promising method for addressing the out-of-distribution ($\displaystyle o.o.d.$) problem, which necessitates bridging the distributional gap between source and target domains. Domain adaptation, a specialized branch of transfer learning spotlighted in studies such as \cite{wilson2020survey} and \cite{patel2015visual}, is particularly aimed at diminishing this $\displaystyle o.o.d.$ discrepancy while maintaining consistent tasks across domains. Our investigation is tailored to domain adaptation scenarios where the source domain is label-rich, in contrast to the target domain where labels are absent \cite{wilson2020survey}.

The trajectory of domain adaptation has seen impressive advancements, predominantly in feature-based transfer learning as per \cite{pan2009survey}. Methods such as Subspace Alignment (SA), detailed in \cite{fernando2013unsupervised}, seek to identify commonalities between the subspaces of the source and target domains via principal component analysis and linear transformations. Additionally, the Optimal Transport for Domain Adaptation (OTDA), introduced by \cite{flamary2016optimal}, proposes a unique domain adaptation methodology based on the optimal transport theory, aiming to find a cost-effective mapping between domains. Substructural Optimal Transport (SOT) \cite{lu2021cross} extends this theory, focusing on the coupling of substructures within the probability distributions. In the realm of deep domain adaptation, the Domain Adversarial Neural Network (DANN) \cite{ganin2016domain} leverages adversarial learning to render a discriminator incapable of distinguishing between the domains, fostering domain-invariant feature creation. The Gaussian-guided feature alignment (GFA) method \cite{zhang2022gaussian} measures and reduces the difference between known features of source subjects and unknown features of target subjects, aligning them for better model adaptation. The Transfer Convolutional (TrC) approach \cite{rokni2018personalized} uses convolutional neural networks to grasp temporal features and then refines the model post-training on the source domain data.

Despite these feature-centric domain adaptation strategies being predominantly applied to static data like images \cite{yang2018learning} \cite{li2020simultaneous}, their application to time series data is becoming increasingly common. However, these techniques may not be fully apt for sensor-based HAR data, as they often overlook the critical temporal dependencies intrinsic to time series data when aligning the distributions. This neglect can lead to a significant loss in the performance of the adapted models. The understanding of temporal relations can illuminate common patterns across users, potentially enriching the domain adaptation process for sensor-based HAR tasks. Therefore, this paper aims to rigorously explore the role of temporal dynamics within the domain adaptation framework, with a particular lens on cross-user HAR in time series data.

\section{Deep Generative Domain Adaptation with Temporal Relation Attention mechanism}

\subsection{Problem formulation}

In a cross-user HAR problem, a labelled source user $\displaystyle S^{Source} =\left\{\left( x_{i}^{Source} ,\ y_{i}^{Source}\right)\right\}_{i=1}^{n^{Source}} $ drawn from a joint probability distribution $\displaystyle P^{Source} $ and a target user $\displaystyle S^{Target} =\left\{\left( x_{i}^{Target} ,\ y_{i}^{Target}\right)\right\}_{i=1}^{n^{Target}} $ drawn from a joint probability distribution $\displaystyle P^{Target} $, where $\displaystyle {n^{Source}} $ and $\displaystyle {n^{Target}} $ are the number of source and target samples respectively.  $\displaystyle S^{Source}$ and $\displaystyle S^{Target}$ have the same feature spaces (i.e. the set of features that describes the data from sensor readings) and label spaces (i.e. the set of activity classes). The source and target users have different distributions, i.e., $\displaystyle P^{Source} \neq P^{Target} $, which means that even for the same activity, the sensor readings look different between the two users. Given source user data $\displaystyle \left\{\left( x_{i}^{Source} ,\ y_{i}^{Source}\right)\right\}_{i=1}^{n^{Source}} $ and target user data $\displaystyle \left\{\left( x_{i}^{Target} \right)\right\}_{i=1}^{n^{Target}} $, the goal is to obtain the labels for the target user activities.

\subsection{An overview of the DGDATA method}

\begin{figure*}[h!]
\centering
\includegraphics[width=\textwidth]{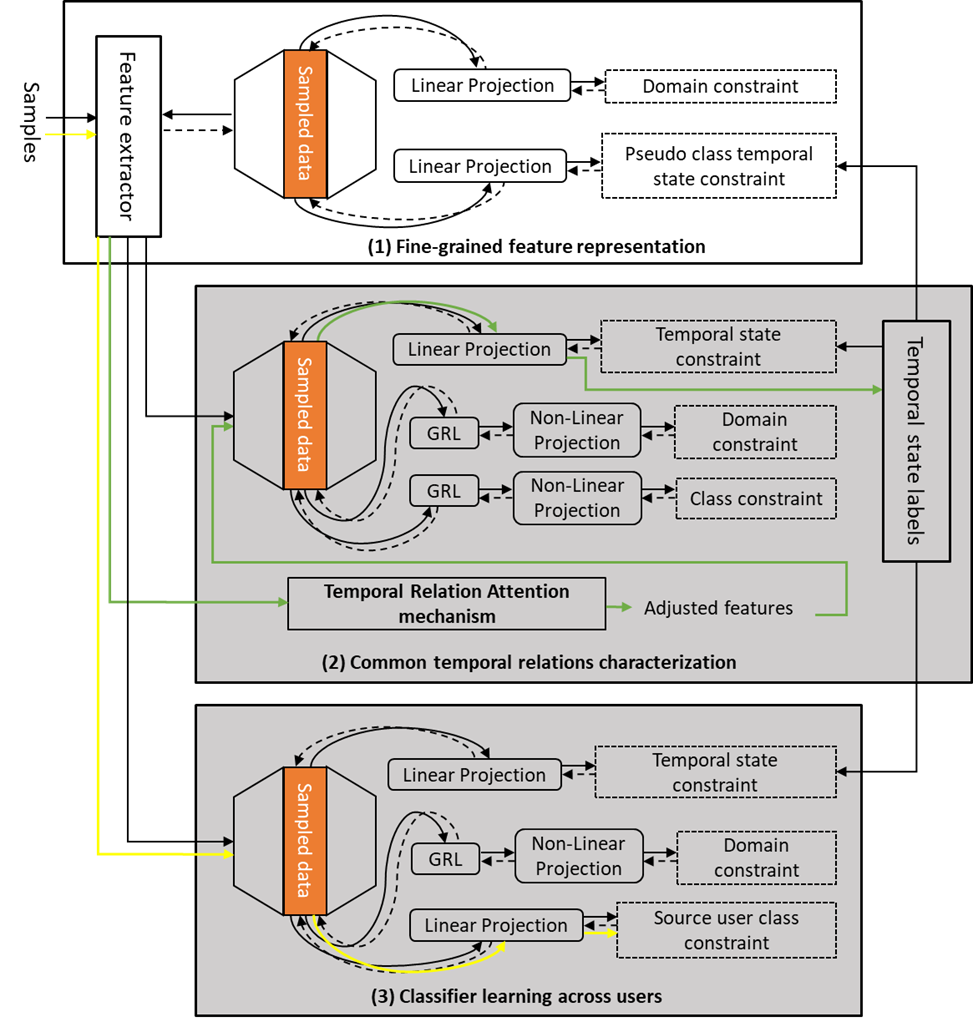}
\caption{The outline of DGDATA.\label{framework_DGDA-TRAM}}
\end{figure*}

In this study, we introduce the Deep Generative Domain Adaptation with Temporal Attention (DGDATA) to tackle the challenges of cross-user HAR. The key observation is that physical movements inherently depend on preceding actions. For instance, the activity of walking can be broken down into three distinct sub-activities: lifting the leg, propelling it forward, and placing the foot on the ground. Our goal is to harness the temporal relations present in human activity time series data, which tend to be consistent across different individuals. This user-invariant temporal knowledge is utilized to enhance the performance and robustness of our proposed domain adaptation model for HAR. The DGDATA method incorporates adversarial learning to leverage these temporal relations for effective cross-user HAR. Specifically, DGDATA comprises three interlinked iterative components, as illustrated in Figure~\ref{framework_DGDA-TRAM}, each addressing a unique aspect of the cross-user HAR challenge.

The components are as follows:

\begin{enumerate}
\item Fine-grained feature representation: This component's role is to discern and extract features that accurately identify temporal states, activity types, and user domains. It serves a similar purpose to a discriminator in Generative Adversarial Networks (GANs) as the white part in Figure~\ref{framework_DGDA-TRAM}, working in conjunction with the other two components (akin to a generator in GANs) to facilitate adversarial learning. This learning process focuses on learning the data distributions of different temporal states, different users and different activity classes. The extracted features are then utilized in the subsequent components.
\item Common temporal relations characterization: The aim here is to identify temporal patterns (i.e. states) that are consistent across different users. It is similar to the role of a generator in GANs, aiming to extract and generate user-invariant temporal state labels that are not distinguishable across users. This is achieved through data distribution mapping in collaboration with the first component. Additionally, the Temporal Relation Attention mechanism is proposed to adjust the distributions of the features. It describes the importance of the previous time steps on the current time step for enhancing the extraction of temporal states and better capturing the temporal relations. Then, the extracted temporal state labels are fed into the other components.
\item Classifier learning across users: This component focuses on using the insights from the previous two to develop a generalized classifier for different users. Similar to a generator in GANs, it aims to learn the activity classifier that is not distinguishable across users. It incorporates the user-invariant temporal relation information from the second component to guide the distribution adjustment, aligning the data distributions between source and target users more effectively. Consequently, a robust classifier is developed, capable of accurately labeling the target user's data.
\end{enumerate}

\begin{figure}[h!]
\centering
\includegraphics[width=0.8\columnwidth]{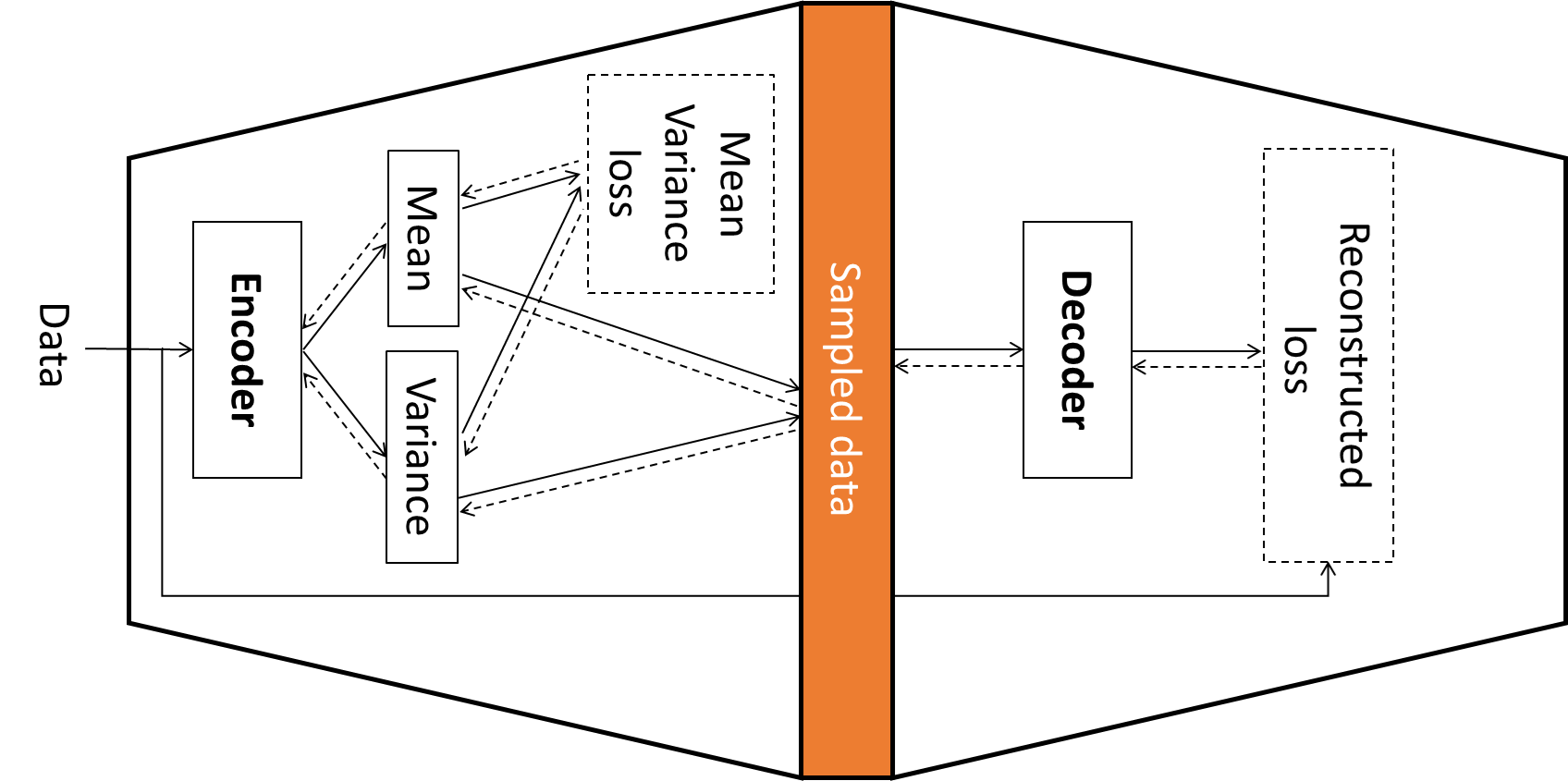}
\caption{The architecture of VAE model.\label{VAE_structure}}
\end{figure}

It is worth mentioning that our method introduces the generative model in Figure~\ref{VAE_structure} as the foundational network architecture applied to the above-mentioned three components for further model generalization improvement. Generative models can generate new data samples similar to the input data for data augmentation and better generalization. Furthermore, even if the labels for some classes are missing during training, generative models try to model the data distribution, which can provide insights into more essential data patterns and structures \cite{ruthotto2021introduction}. As for our cross-user HAR task, generative models can be more suitable for modeling the user-invariant activity data distributions considering the unavailability of target user labels. Because generative models encode data into a compact latent space, this compact representation can capture the underlying factors or patterns in the data \cite{van2010activity}. When a classification task transfers from one user to another user, if both users share some common underlying patterns, then this latent space can be shared across the users.

In the proposed approach, we consider the Variational Autoencoder (VAE) \cite{Kingma2014} architecture as the foundational model. VAE strives to learn a structured latent space where similar data samples are close together such as patterns in time series data, and this space generally follows a Gaussian distribution. As in Figure~\ref{VAE_structure}, VAE is a generative model that takes input data (e.g., sensor features in HAR) and compresses it into a lower-dimensional latent space using an encoder. This latent space is represented probabilistically with mean and variance parameters, encapsulating the intrinsic characteristics in human activities or time series trends. To ensure differentiability during training, the VAE employs the reparameterization trick, where it samples from the latent space by sampling from a Gaussian distribution, scaling by the learned variance and shifting by the learned mean. The sampled latent representation is then passed through a decoder to reconstruct the original input. This approach allows for a nuanced understanding and generation of complex activity patterns, making VAE an ideal model for our cross-user HAR task.

\begin{figure*}[h!]
\centering
\includegraphics[width=1.1\textwidth]{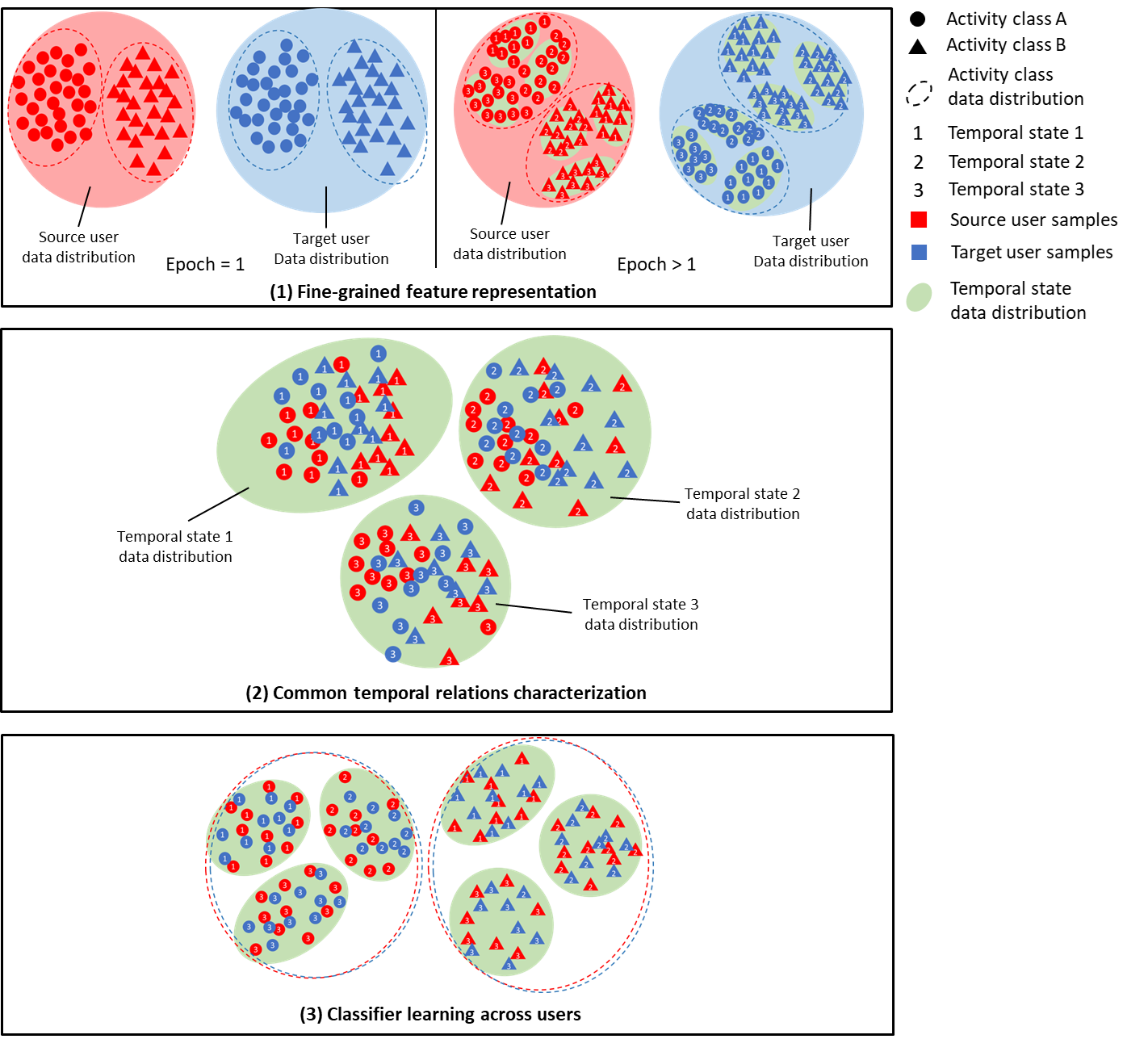}
\caption{The outline of DGDATA data distributions mapping.\label{framework_DGDA-TRAM-intuition}}
\end{figure*}

This sampling process of VAE has several benefits regarding generalization and domain adaptation. The injection of randomness due to sampling acts as a form of regularization. Instead of directly encoding the data into a deterministic latent point, the VAE encodes it into a distribution. This prevents the model from overfitting to the training data, allowing for better generalization. By sampling from a distribution and enforcing the latent variables to be close to a Gaussian distribution through the Kullback–Leibler (KL) divergence term, the latent space tends to be more structured and continuous \cite{rybkin2021simple}. When source and target domains are represented in such a space, alignment techniques such as the Gradient Reversal Layer (GRL) adversarial learning strategy that will be discussed in the following section can be more effectively applied.

Overall, DGDATA is designed to bridge the gap between source and target users by capitalizing on temporal relation information. The three components function in a sequential and synergistic manner. The first component identifies specific features that differentiate users, activities, and temporal states. The second component collaborates with the first by extracting the common temporal relations across users, focusing on similarities rather than differences. Moreover, the temporal state labels learned in the second component inform and refine the feature extraction in the first component during subsequent iterations. The third component synthesizes the differences highlighted by the first component and the similarities from the second to construct a classifier that is both robust and generalizable. This method can be seen as a form of adversarial learning, playing on the differences and similarities in the data, and further improving the alignment between source and target domain distributions through temporal relation information and generative learning. The following three subsections will detail the design principles and implementation of each component.

\subsection{Fine-grained feature representation}

In this section, we focus on developing a unique feature extractor that is central to our DGDATA method. This extractor is designed to extract features to discern intricate data distributions, distinguishing between various users, activity categories, and temporal states. Such detailed feature extraction lays the groundwork for the subsequent components of our method. It provides the necessary information for the other two components to effectively extract user-independent temporal states and to develop a classifier that can generalize across different users.

Figure~\ref{framework_DGDA-TRAM-intuition} offers an intuitive visual representation of the data distribution adjustment for each component. This diagram is meant to give a clear picture of how the data distributions are adjusted and aligned throughout the process. In Figure~\ref{framework_DGDA-TRAM-intuition} (1), there are two parts: the initial setup at the start of the model training (left) and the evolving distributions in later training epochs (right). In these illustrations, the source user data is marked in red, and the target user data is in blue. Different shapes indicate various activity classes.

Conditional Variational Autoencoder (CVAE) \cite{kim2021conditional} is a type of VAE that incorporates conditional variables as a guide for learning data distribution generation. In this way, CVAE allows the additional constraint conditions of the domain, activity classes and the corresponding temporal states information for improving the model's discriminative capability under the data distribution generation process \cite{lasserre2006principled}. There are two constraints, and the first is called domain constraint. This constraint focuses on making the learned data distribution more distinguishable between the source and target users; as it can be seen in Figure~\ref{framework_DGDA-TRAM-intuition} (1), the data distributions between the source and target users become more different in later iterations. Domain labels for this constraint are provided, facilitating a binary classification to differentiate between the source and target users. The other constraint is called the pseudo class temporal state constraint, and is designed to distinguish between various activity classes and their associated temporal states (as the red and blue dashed ellipses, and the green filled ellipses shown in Figure~\ref{framework_DGDA-TRAM-intuition} (1). For this specific constraint, activity labels for the source user are accessible, but those for the target user remain unavailable. 

In the first training epoch, temporal states haven’t yet been determined in component (2), so we initially set all temporal state labels to 0 (i.e., $\displaystyle ts= {0,0,...,0}$). As the model progresses and temporal state labels $\displaystyle ts$ become available from component (2), these labels are used in feature extraction. In Figure~\ref{framework_DGDA-TRAM-intuition} (1), the green-filled ellipses indicate distributions that separate different temporal states within each activity class. For example, the numbers one, two and three in triangles correspond to raising the leg, thrusting forward and feet to the ground of walking activity.

\begin{figure}[h!]
\centering
\includegraphics[width=0.7\textwidth]{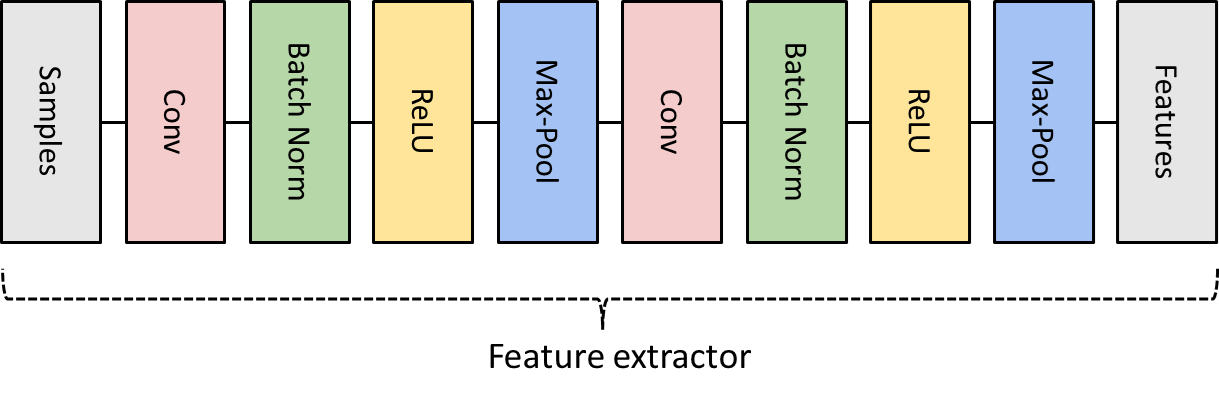}
\caption{The network architecture of feature extractor.\label{feature_extractor}}
\end{figure}

In the proposed network architecture, as depicted in Figure~\ref{feature_extractor}, the feature extraction encompasses two primary convolutional layers. Each of these layers incorporates a 1D convolutional layer tailored specifically for extracting temporal features from the time series data. Subsequent to this convolution operation, batch normalization is applied to mitigate the internal covariate shift, enhancing model stability and convergence. This is then followed by the application of a Rectified Linear Unit (ReLU) activation function, facilitating the introduction of non-linear transformations. Lastly, a max-pooling procedure is implemented to execute downsampling. Upon channeling the input data samples through these consecutive layers, the refined and learned features are consequently produced.

\begin{figure}[h!]
\centering
\includegraphics[width=0.8\columnwidth]{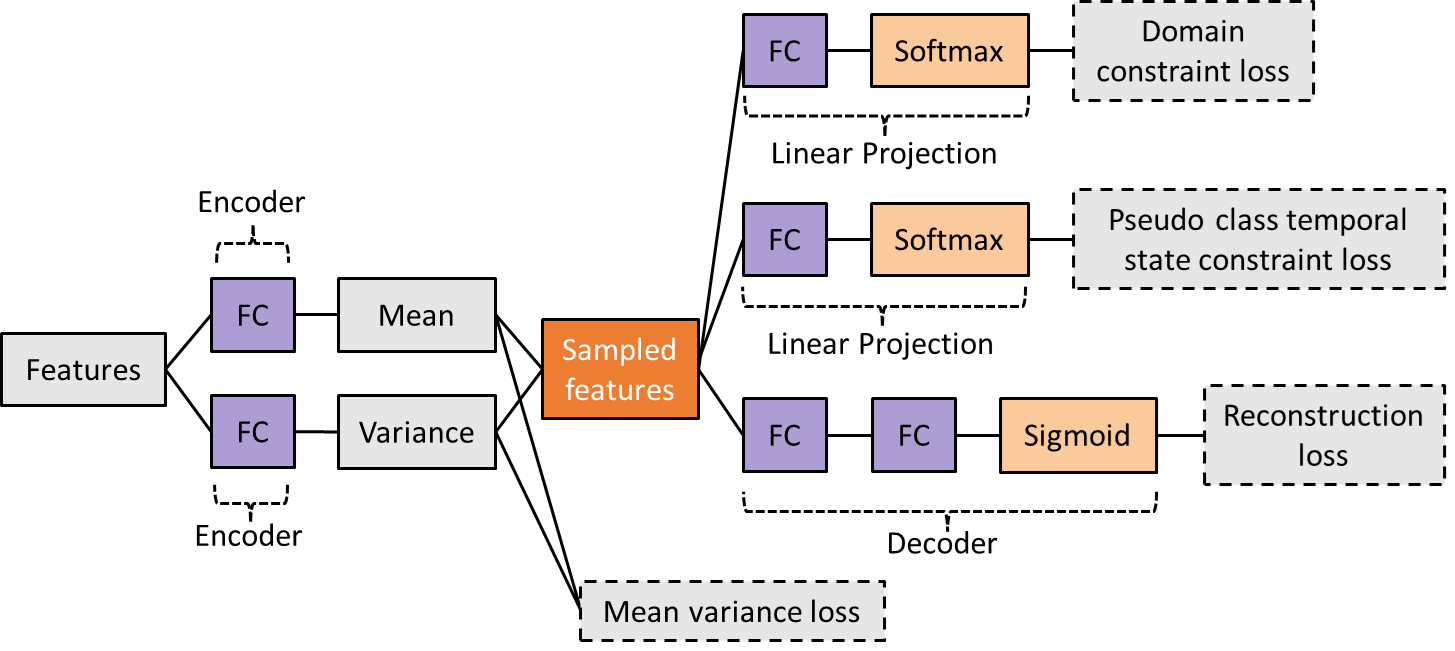}
\caption{The network architecture of fine-grained feature representation component.\label{Fine_grained_feature_representation}}
\end{figure}

The network architecture of the fine-grained feature representation component is shown in Figure~\ref{Fine_grained_feature_representation}. The architecture integrates a CVAE with two dedicated constraints, i.e. domain constraint and pseudo class temporal state constraint. The network architecture begins by taking input features and passing them through an encoder that uses two linear layers to condense the features into mean and variance of Gaussian distributions. This is followed by a sampling phase, injecting randomness to ensure robustness and generalization. The decoder then attempts to reconstruct the original input through successive linear transformations. Then, a sigmoid function is applied to make the features in a normalization way. By ensuring that the model sees features on the same scale during training, normalization can help the model generalize better to unseen data. Simultaneously, domain constraint and pseudo class temporal state constraint analyse the sampled features, facilitating both activity multi-class and domain binary categorizations anchored on the data's fundamental characteristics. 

Here, we explain the design of loss functions. $\displaystyle h_{f} $ is the feature extractor for the global framework. $\displaystyle h_{emf}$, $\displaystyle h_{evf}$, $\displaystyle h_{df}$, $\displaystyle h_{dcf}$, and $\displaystyle h_{pctscf}$ are encoder mean, encoder variance, decoder, domain constraint and pseudo class temporal state constraint of fine-grained feature representation component. 

\text{Pseudo Class Temporal State Constraint Loss:}
\begin{equation}
L_{pctscf} = l_{CE}\big( h_{pctscf}(S(h_{emf}(h_{f}(x)), h_{evf}(h_{f}(x))), \hat{y} \big)
\end{equation}

\text{Domain Constraint Loss:}
\begin{equation}
L_{dcf} = l_{CE}\big( h_{dcf}(S(h_{emf}(h_{f}(x)), h_{evf}(h_{f}(x))), d \big)
\end{equation}

where $\displaystyle l_{CE}$ is the cross-entropy loss function, the domain labels are $d \in \{0,1\}$. $S()$ symbolizes the sampling phase.

In VAEs, the KL divergence term is added to enforce that the latent space distribution resembles a specific distribution (usually a standard normal distribution). We modify the KL divergence term to give the latent space a given variance $\displaystyle var$ instead of the standard normal distribution's variance, considering the latent space's unknown data distribution structural properties. The mean constraint is still centred around zero. This is a common setting in VAEs to ensure the latent space doesn't drift too far from a known structure. Therefore, the loss functions are designed as below.

\text{Mean Variance Loss:}
\begin{equation}
L_{mvf} = l_{MSE}\big( h_{emf}(h_{f}(x)), 0 \big) + l_{KL}\big( h_{evf}(h_{f}(x)), \text{var} \big)
\end{equation}

\text{Decoder Reconstruction Loss:}
\begin{equation}
L_{reconf} = l_{MSE}\big( h_{df}(S( h_{emf}(h_{f}(x)), h_{evf}(h_{f}(x))), h_{f}(x) \big)
\end{equation}

\text{The Total Loss:}
\begin{equation}
L_{f} = \alpha*L_{reconf} + \zeta*L_{mvf} + \gamma*L_{pctscf} + \delta*L_{dcf}
\end{equation}

Where $\displaystyle l_{MSE}$ is the mean squared error loss function. $\displaystyle l_{KL}$ is the the KL divergence loss function. $\displaystyle \alpha, \zeta, \gamma, \delta$ are the weight coefficients of different loss functions.

\subsection{Common temporal relations characterization}

\subsubsection{The overarching workflow}

This component is designed to identify common temporal relations of activities over time, which are shared across various users. While individuals may perform these activities in their own unique way, the sequences of these sub-activities are the same across different people due to the temporal dependency of physical movements. This component focuses on extracting these sub-activities, identifying the variations within them, and outlining the common sequences of activities that occur across different users. Essentially, it finds and describes the shared temporal relations of the sub-activities within activities, regardless of individual differences in how these activities are carried out.

\begin{figure*}[h!]
\centering
\includegraphics[width=\textwidth]{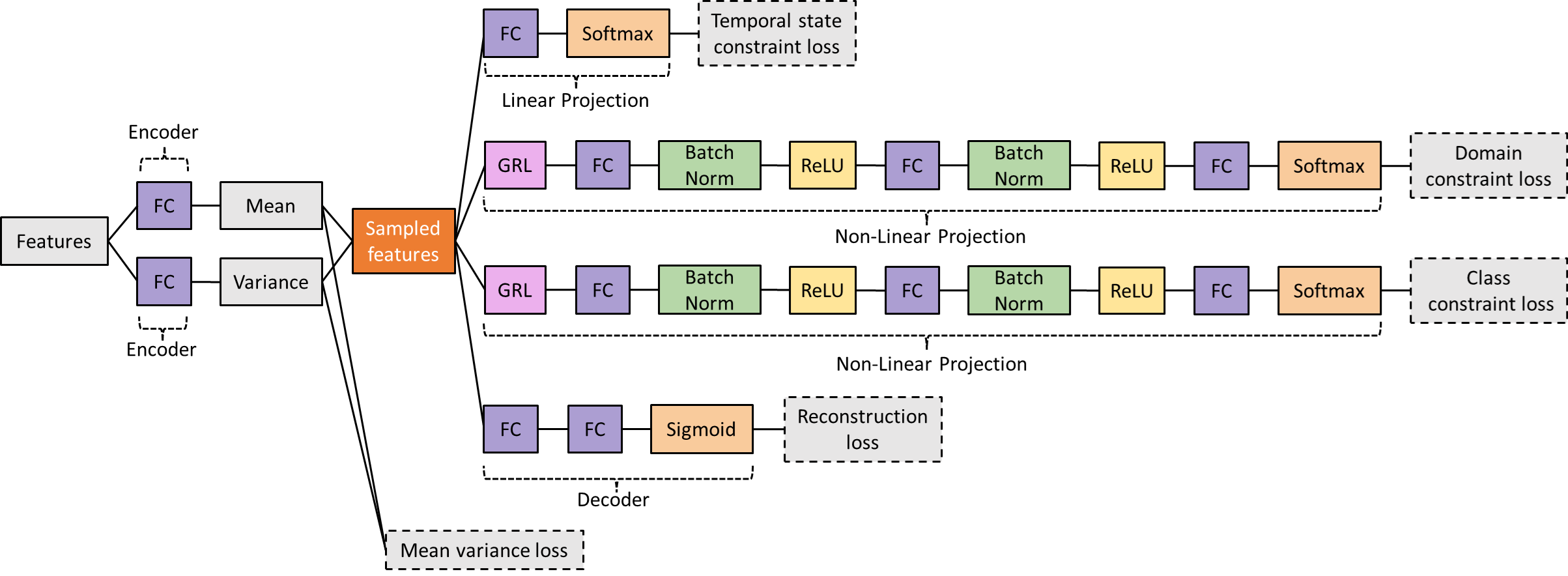}
\caption{The network architecture of common temporal relations characterization component.\label{latent_temporal_states_characterization}}
\end{figure*}

The synergy between this component and the first one is akin to the interplay between discriminator and generator in the realm of GANs applied to domain adaptation. In this analogy, the discriminator (akin to the first component) is trained to recognize the differences in data distributions across users and activities. In contrast, the generator (analogous to this component) strives to blur these distinctions, making the data distributions increasingly similar. This adversarial dynamic is essential for refining the feature extraction in the first component and for aligning the data distributions. It's a continual tug-of-war that underpins the enhanced representation of activity classes and users in the overall learning process.

This component uncovers essential patterns in human activities, providing a foundational understanding that is not affected by individual differences. This knowledge of consistent temporal relations serves as a guide for the other two components. Particularly, when the third component develops a classifier, this foundational insight helps ensure that the classifier remains broadly applicable and doesn't become overly specialized to the unique behaviors of a single user. This approach improves the classifier's ability to be applied effectively across different individuals.

Within this component itself, adversarial learning strategy \cite{ganin2016domain} is also applied. This strategy updates the loss of temporal state constraint and uses gradient reversal layer (GRL) \cite{ganin2016domain} which is an adversarial learning technique via reversing gradients to update the loss of class constraint and domain constraint. This design ensures that the derived features are not merely specific to the temporal state but also exhibit a generalized nature. This generalization reduces their sensitivity to their originating domains and activity classes for obtaining user-invariant class-invariant temporal state representations. As in Figure~\ref{framework_DGDA-TRAM-intuition} (2), the green filled ellipses are the different data distributions across different temporal states. The number in each shape indicates the corresponding temporal state information hidden in the time series data.

In Figure~\ref{latent_temporal_states_characterization}, temporal state constraint uses a linear transformation to guide the data distribution generation based on temporal state categories. Concurrently, the model incorporates two other constraints: i.e. domain and class constraints. Domain constraint aims to discern the users from which they originate, while class constraint focuses on identifying the different activity classes. Central to these two constraints is the introduction of the GRL which is used to blur the data distributions of the users and activity classes. These two constraints' network architectures are the same, consisting of multiple fully connected layers enhanced with batch normalization, ReLU activation function and a softmax output layer. The multi-layered design, fortified with batch normalization and nonlinear activations, makes the network more complex. This is crucial as the adversarial learning strategy demands that these two constraints consistently modify and refine their data distributions to effectively learn user-invariant temporal states, often necessitating a sophisticated architecture for optimal performance.

We leverage a series of loss functions for the common temporal relations characterization component. These loss functions are from several core modules of the network architecture including encoder mean ($h_{emt}$), encoder variance ($h_{evt}$), decoder ($h_{dt}$), domain constraint ($h_{dct}$), temporal state constraint ($h_{tct}$), and class constraint ($h_{cct}$). Here we detail each loss function:

\text{Temporal State Constraint Loss:}
\begin{equation}
L_{tsct} = l_{CE}\big( h_{tct}(S(h_{emt}(h_{f}(x)), h_{evt}(h_{f}(x))), \hat{ts} \big) 
\end{equation}

\text{Domain Constraint Loss:}
\begin{equation}
L_{dct} = l_{CE}\big( h_{dct}(R(S(h_{emt}(h_{f}(x)), h_{evt}(h_{f}(x)))), d \big)
\end{equation}

\text{Class Constraint Loss:}
\begin{equation}
L_{cct} = l_{CE}\big( h_{cct}(R(S(h_{emt}(h_{f}(x)), h_{evt}(h_{f}(x)))), c \big)
\end{equation}

\text{Mean Variance Loss:}
\begin{equation}
L_{mvt} = l_{MSE}\big( h_{emt}(h_{f}(x)), 0 \big) + l_{KL}\big( h_{evt}(h_{f}(x)), \text{var} \big)
\end{equation}

\text{Decoder Reconstruction Loss:}
\begin{equation}
L_{recont} = l_{MSE}\big( h_{dt}(S(h_{emt}(h_{f}(x)), h_{evt}(h_{f}(x)))), h_{f}(x) \big)
\end{equation}

\text{The Toal Loss:}
\begin{equation}
L_{t} = \alpha*L_{recont} + \zeta*L_{mvt} + \gamma*L_{cct} + \delta*L_{dct} + \eta*L_{tsct} 
\end{equation}

Where $\displaystyle l_{CE}$ is the cross-entropy loss function, $\displaystyle \hat{ts}$ is the pseudo temporal state label (elaborated in subsequent sections). $R$ represents the gradient reversal layer. $\displaystyle \alpha, \zeta, \gamma, \delta, \eta$ are the weight coefficients of different loss functions.

\subsubsection{Temporal Relation Attention mechanism}

For the temporal states $\displaystyle \hat{ts} $ annotation, Temporal Relation Attention mechanism considering user-invariant temporal relation is proposed. The green arrows shown in Figure~\ref{framework_DGDA-TRAM} (2) are the process of the Temporal Relation Attention mechanism. Temporal relation is the essential property of time series data, and different users’ time series activity data follow the same temporal relation in various activities. To achieve correct temporal state labels, not only the feature vectors' distance but also the feature vectors from adjacent past time points should be considered. This idea acknowledges that a feature vector is not only defined by its current time point but also by how it arrived at this time point from the previous several time points. Incorporating this temporal relation context is essential for a more holistic understanding of the time series data. By modelling the sequential nature of the data, we gain insights into the underlying patterns and trends that are not immediately apparent when examining individual feature vectors in isolation. This method provides a richer, more nuanced understanding of the time series data across users, ensuring that the common temporal states are accurately labelled and truly reflective of the underlying dynamics of the activity data across users.

To capture the temporal relation across the samples in sequence and represent the different temporal relation importance of past neighboring samples, regression is applied to the features from both source and target users for obtaining user-invariant temporal patterns in a self-supervised learning way. Regression is a type of supervised learning where the goal is to predict a continuous output variable based on one or more input variables. The multiple Linear Regression model extends the simple linear regression to include multiple predictor variables and it is suitable for our task. The expression is shown below:

\begin{equation}
h_{f}(x_t) = \beta_1 h_{f}(x_{t-1}) + \beta_2 h_{f}(x_{t-2}) + \dots + \beta_p h_{f}(x_{t-p})
\end{equation}

$\displaystyle h_{f}(x_t) $ is the feature in the current time point $\displaystyle t $, $\displaystyle h_{f}(x_{t-1}), h_{f}(x_{t-2}), \dots, h_{f}(x_{t-p}) $ are the features from previous $\displaystyle p$ time points, and $\displaystyle \bm{\beta}={\beta_1, \beta_2, \dots, \beta_p} $ are the weights of the $\displaystyle p$ features.

Given a specific number of time lags, denoted as $\displaystyle p$, the set of weights $\displaystyle \bm{\beta}$ are determined through the regression process. Each weight in this set represents the influence of a feature from a past time point on the current feature value. The value of each weight indicates how strong this influence is. A larger weight implies a greater impact of the corresponding past feature on the current feature.

\begin{figure}[h!]
\centering
\includegraphics[width=0.5\columnwidth]{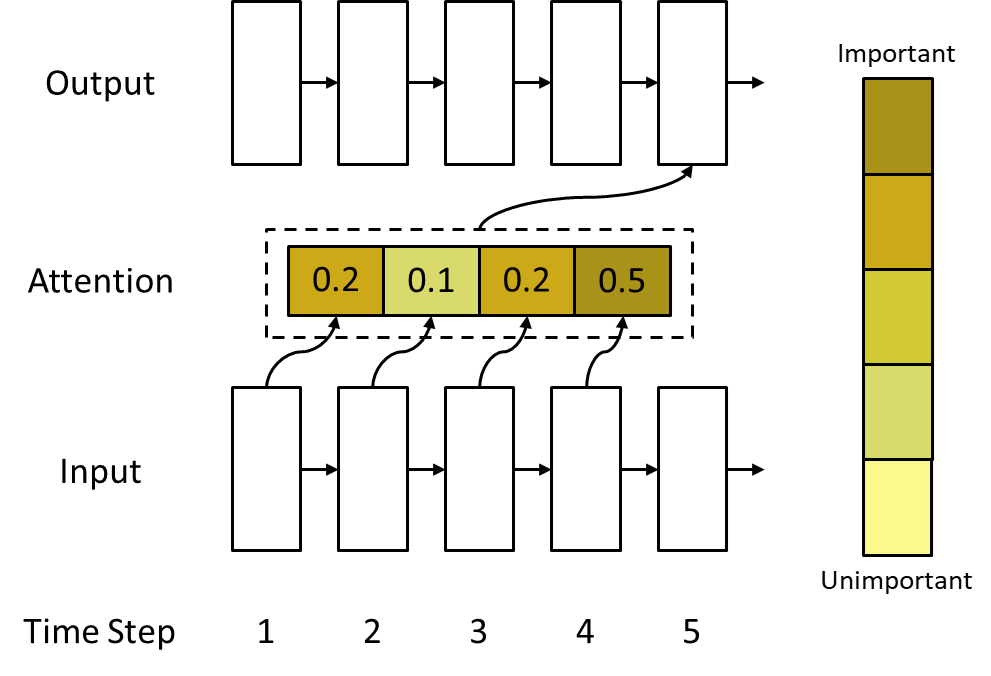}
\caption{Temporal Attention.\label{attention_time_series}}
\end{figure}

The weights in our model are similar to the attention weights found in attention mechanisms. In attention mechanisms, these weights determine how much emphasis the model should give to certain parts of the input. In our context, the weights indicate the significance of features from past time points when calculating the feature's current value. For instance, when the number of time lags is set to 4, this relationship is visually depicted in Figure~\ref{attention_time_series}. This figure helps illustrate how features from the four previous time points influence the current feature's value.

After the initial feature extraction process, temporal attention is applied. This process involves combining the current feature vector with several of the most significant past feature vectors. By doing this, we 'correct' or refine the current feature, taking into account the most relevant information from its recent history. This approach enhances the accuracy and relevance of the feature representation at the current moment.

\subsection{Classifier learning across users}

This component's goal is to develop a classifier for user-independent activity recognition, enhancing the effectiveness of cross-user HAR. The user-independent temporal states, obtained from the second component, serve as a unified representation to bridge the variability across different users. These temporal states provide essential prior knowledge for the classifier that is meant to work across different users. Since these temporal states are not specific to any one user, they offer generalized activity sequences that apply to a wide range of users. Acting as a form of regularization, these temporal states help prevent the classifier from becoming too narrowly focused on the characteristics of individual users. This component helps in constraining the HAR process to be more user-agnostic.

\begin{figure*}[h!]
\centering
\includegraphics[width=\textwidth]{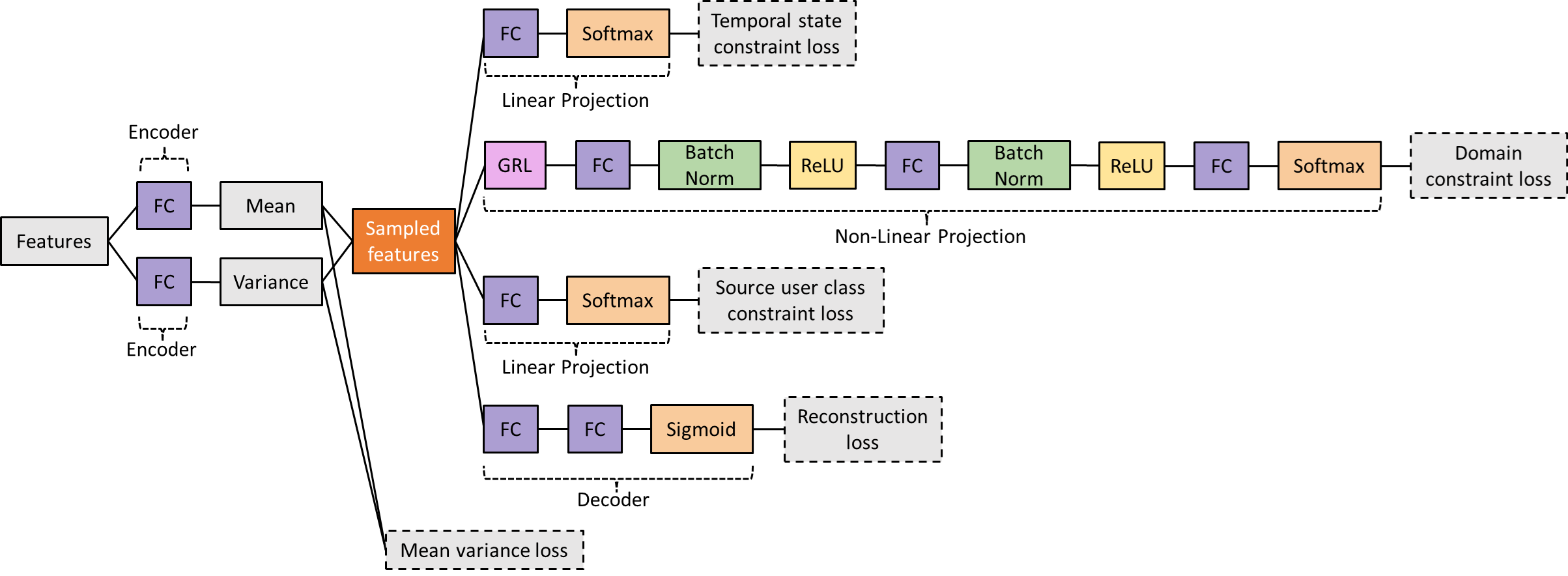}
\caption{The network architecture of cross-user classifier learning component.\label{cross_user_classifier_learning}}
\end{figure*}

Additionally, this component engages in adversarial learning with component (1). While component (1) functions similarly as a discriminator in GANs, distinguishing between different users based on sensor data, this component operates like a GAN generator, aiming to blur the distinctions between source and target user data distributions. Concurrently, it also focuses on recognizing activity labels from the source user. This iterative interaction between components (1) and (3) leads to continual refinement and improvement, culminating in a robust and effective user-invariant activity classification system.

Within this component, we employ the GRL technique for adversarial learning. This method helps in adjusting the feature distribution to not only maintain user-invariant temporal relation knowledge but also to ensure that the data distributions of activity classes are applicable across different users as shown in Figure~\ref{framework_DGDA-TRAM-intuition} (3). Figure~\ref{framework_DGDA-TRAM} (3) illustrates the detailed design, where the temporal state constraint is designed to accurately learn the data distributions of temporal states, and the source user class constraint is focused on activity classification. Simultaneously, the domain constraint aims to confuse the domain discriminators. This confusion is pivotal in driving the learning process towards learning user-invariant data distributions for activity classes.

In Figure~\ref{cross_user_classifier_learning}, after acquiring the sampled features, they are also fed into the source user class constraint. This constraint employs a linear transformation to guide the generation of data distribution, taking into account the activity class labels of the source user. For the design of the loss functions, $\displaystyle h_{emc}$, $\displaystyle h_{evc}$, $\displaystyle h_{dc}$, $\displaystyle h_{dcc}$, $\displaystyle h_{tcc}$, and $\displaystyle h_{succc}$ are encoder mean, encoder variance, decoder, domain constraint, temporal state constraint and source user class constraint of classifier learning across users component respectively. The loss functions are defined as follows:

\text{Temporal State Constraint Loss:}
\begin{equation}
L_{tscc} = l_{CE}\big( h_{tcc}(S(h_{emc}(h_{f}(x)), h_{evc}(h_{f}(x))), \hat{ts} \big)
\end{equation}

\text{Domain Constraint Loss:}
\begin{equation}
L_{dcc} = l_{CE}\big( h_{dcc}(R_{\lambda}(S(h_{emc}(h_{f}(x)), h_{evc}(h_{f}(x)))), d \big)
\end{equation}

\text{Source User Class Constraint Loss:}
\begin{equation}
\begin{split}
L_{succc} = & l_{CE}\big( h_{succc}(S(h_{emc}(h_{f}(x_{source})), \\ 
& h_{evc}(h_{f}(x_{source}))), c_{source} \big)
\end{split}
\end{equation}

\text{Mean Variance Loss:}
\begin{equation}
L_{mvc} = l_{MSE}\big( h_{emc}(h_{f}(x)), 0 \big) + l_{KL}\big( h_{evc}(h_{f}(x)), \text{var} \big)
\end{equation}

\text{Decoder Reconstruction Loss:}
\begin{equation}
L_{reconc} = l_{MSE}\big( h_{dc}(S(h_{emc}(h_{f}(x)), h_{evc}(h_{f}(x)))), h_{f}(x) \big)
\end{equation}

\text{The Total Loss:}
\begin{equation}
L_{c} = \alpha*L_{reconc} + \zeta*L_{mvc} + \gamma*L_{succc} + \delta*L_{dcc} + \eta*L_{tscc}
\end{equation}

Where $\displaystyle \alpha, \zeta, \gamma, \delta, \eta$ are the weight coefficients of different loss functions.

\begin{figure}[h!]
\centering
\includegraphics[width=0.6\columnwidth]{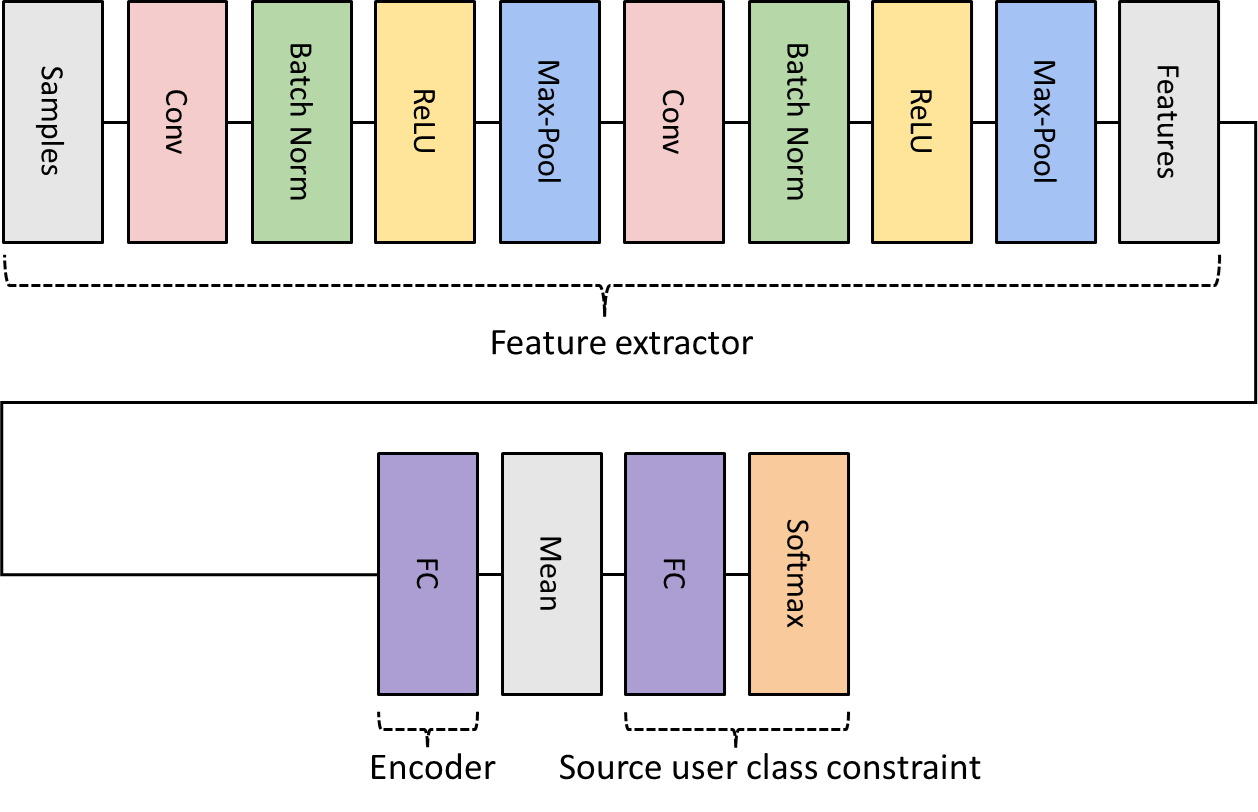}
\caption{The network architecture of target user classification.\label{target_classification}}
\end{figure}

The process of the DGDATA method that iteratively trains the three components. After training, the encoder and the source user class constraint components in the classifier learning across users component combined with the global feature extractor is the final model we need for target user classification as shown in Figure~\ref{target_classification}. The process of classification inference is shown as the yellow arrows in Figure~\ref{framework_DGDA-TRAM} (3).

\section{Experiments}

In this section, we assess the effectiveness of the proposed DGDATA method through comprehensive experiments on cross-user HAR.

\subsection{Datasets and Experimental setup}
In this study, we employ three publicly available sensor-based human activity recognition (HAR) datasets to validate our cross-user HAR methodology. A summary of the selected subjects and common activities across datasets is provided in Table~\ref{tab_datasets_info}. For this investigation, we restrict our focus to accelerometer and gyroscope sensor data acquired from the lower right arm, simulating a smartwatch-based use-case scenario.

\begin{table}[h!]
\caption{Three sensor-based HAR datasets information.}
\label{tab_datasets_info}
\centering
\resizebox{\textwidth}{!}{%
\begin{tabular}{|p{1.5cm}|p{1.5cm}|p{2cm}|p{8.7cm}|}
\hline
\textbf{Dataset} & \textbf{Subjects} & \textbf{\#Activities} & \textbf{Common Activities} \\ \hline
OPPT & S1, S2, S3 & 4 & 1 standing, 2 walking, 3 sitting, 4 lying \\ \hline
PAMAP2 & 1, 5, 6 & 11 & \begin{tabular}[c]{@{}l@{}}1 lying, 2 sitting, 3 standing, 4 walking, \\ 5 running, 6 cycling, 7 Nordic walking, \\ 8 ascending stairs, 9 descending stairs, \\ 10 vacuum cleaning, 11 ironing\end{tabular} \\ \hline
DSADS & 2, 4, 7 & 19 & \begin{tabular}[c]{@{}l@{}}1 sitting, 2 standing, 3 lying on back, \\ 4 lying on right, 5 ascending stairs, \\ 6 descending stairs, 7 standing in elevator still, \\ 8 moving around in elevator, \\9 walking in parking lot, \\10 walking on treadmill in flat, \\11 walking on treadmill inclined positions, \\12 running on treadmill in flat, \\13 exercising on stepper, \\14 exercising on cross trainer, \\15 cycling on exercise bike in horizontal positions, \\16 cycling on exercise bike in vertical positions, \\17 rowing, 18 jumping, 19 playing basketball\end{tabular} \\ \hline
\end{tabular}%
}
\end{table}

The OPPORTUNITY (OPPT) dataset \cite{chavarriaga2013opportunity} offers recordings of subjects engaged in everyday morning activities. The subjects carry out activities without any imposed limitations, following only a generalized description of the tasks to be performed. The PAMAP2 physical activity monitoring dataset \cite{reiss2012introducing} adheres to a more structured protocol, specifying particular activities for each participant, with over 10 hours of cumulative data collected. Lastly, the Daily and Sports Activities Data Set (DSADS) \cite{barshan2014recognizing} is gathered by instructing participants to perform activities according to their own style. This lack of uniform guidelines may contribute to an increased level of inter-subject variability, better mimicking real-world conditions. Each activity is executed by each participant for a duration of 5 minutes.

These datasets are chosen based on ascending levels of domain adaptation complexity. Specifically, OPPT contains 4 distinct activities, PAMAP2 with 11 activities, and DSADS with 19 activities. Notably, the DSADS dataset presents the added challenge of classifying closely related activities, such as three different forms of walking, rendering it more complex to differentiate.

For the experimental setup of cross-user HAR, we utilize the sliding window technique, a commonly employed data segmentation approach in sensor-based HAR \cite{wang2018impact}. To capture temporal relation, each window is configured to have a fixed time interval of 3 seconds, with a 50\% overlap, consistent with the common setting in sensor-based HAR tasks.

We employ six methods for comparison in the categories of traditional domain adaptation and deep domain adaptation methods. Notably, TrC requires a small amount of labels from the target domain for fine-tuning, whereas the other methods do not.

\textbf{Traditional domain adaptation:}

\begin{enumerate}
\item \textbf{SA} \cite{fernando2013unsupervised}: This approach aims to align the subspaces between source and target domains to make them more similar in the subspace.

\item \textbf{OTDA} \cite{flamary2016optimal}: Based on optimal transport theory, this method seeks to transform one probability distribution into another in the most efficient way, bridging the source and target domain samples.

\item \textbf{CORAL} \cite{sun2016return}: CORAL aligns the covariance of feature layers, promoting better domain-agnostic features.

\item \textbf{SOT} \cite{lu2021cross}: This method delves into the underlying structure of domains to facilitate mapping at the substructure level. It strikes a balance between broad and detailed mapping.
\end{enumerate}

\textbf{Deep domain adaptation:}

\begin{enumerate}
\item \textbf{DANN} \cite{ganin2016domain}: A deep domain adaptation approach that uses adversarial training. Its goal is to make a discriminator unable to differentiate between domains, thus promoting features that are consistent across domains.
\item \textbf{TrC} \cite{rokni2018personalized}: This is a deep domain adaptation strategy that utilizes CNNs to discern temporal attributes. After training on data from the source domain, the model undergoes fine-tuning in the target domain.
\end{enumerate}

\begin{table}[h]
\centering
\begin{tabular}{|l|l|}
\hline
\textbf{Parameter} & \textbf{Value} \\
\hline
Training Epochs &  100 \\
Adam Optimizer Weight Decay & 0.0005 \\
Adam Optimizer Beta & 0.2 \\
Reconstruction Loss Coefficient $\alpha$ & 1.0 \\
Mean-Variance Loss Coefficient $\zeta$ & 10.0 \\
Class Constraint Loss Coefficient $\gamma$ & 30.0 \\
Domain Constraint Loss Coefficient $\delta$ & 1.0 \\
Temporal State Constraint Loss Coefficient $\eta$ & 10.0 \\
\hline
\end{tabular}
\caption{Parameter settings of the DGDATA Method.\label{DGDATA_paras}}
\end{table}

For these methods, we adjust hyper-parameters, taking a cue from the methodology \cite{flamary2016optimal} to avoid overfitting during tests. We split the target user data into validation and test subsets. The validation subset helps fine-tune hyper-parameters for best accuracy. Once these hyper-parameters are finalized, we test the model's efficacy on the test subset. The primary metric for assessment is the classification accuracy for the target user. Regarding the DGDATA method, parameter settings are listed in Table~\ref{DGDATA_paras}. The different coefficients of constraint loss are set based on the importance of these constraints.

\subsection{Cross-user HAR performance evaluation}
We begin by examining the classification outcomes from various methods. These evaluations are performed on the OPPT, PAMAP2, and DSADS datasets. Each method's efficacy is assessed based on its ability to transition between individual users. Within each dataset, three users are chosen at random from the entire user pool. Subsequently, a one-to-one cross-user HAR task is executed, covering every potential user pairing within each dataset.

\begin{figure}[h!]
\centering
\includegraphics[width=\columnwidth]{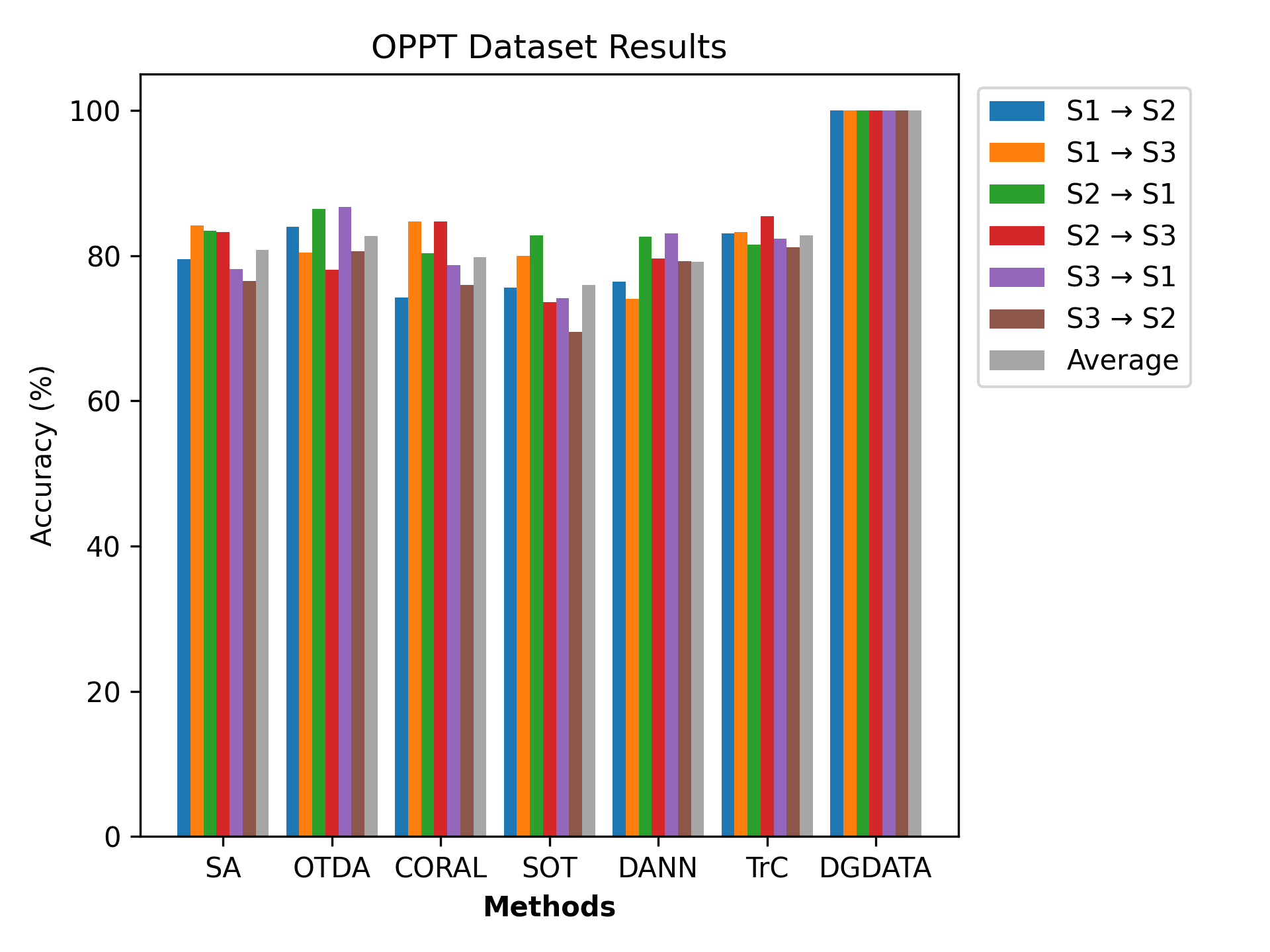}
\caption{OPPT dataset classification results.\label{OPPT_Dataset_Results}}
\end{figure}

On the OPPT dataset (see Figure~\ref{OPPT_Dataset_Results}), DGDATA consistently outperforms other methods, achieving a perfect score of 100\% in all test scenarios, indicating that this model is highly effective and adaptable across varying conditions in the OPPT dataset. TrC also performs relatively well, with accuracy consistently in the low to mid-80\%, signifying its competence but not matching the excellence of DGDATA. SA, OTDA, and DANN exhibit moderate performance, with accuracy mostly fluctuating between mid-70\% to mid-80\%. SOT and CORAL display slightly lower performances.

\begin{figure}[h!]
\centering
\includegraphics[width=\columnwidth]{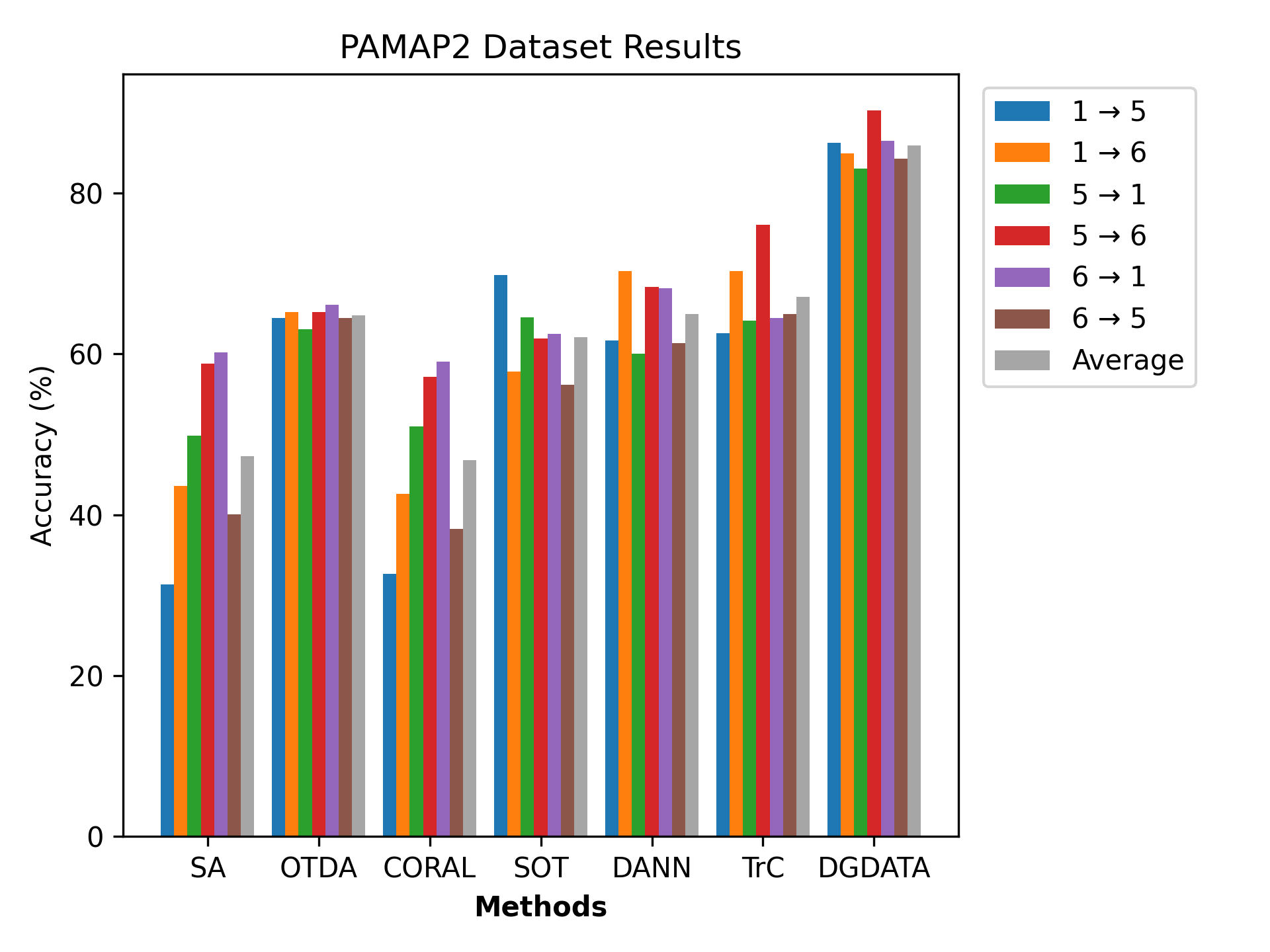}
\caption{PAMAP2 dataset classification results.\label{PAMAP2_Dataset_Results}}
\end{figure}

On the PAMAP2 dataset (see Figure~\ref{PAMAP2_Dataset_Results}), DGDATA dominates, with accuracy above 83\%, peaking at 90.29\% in the 5 $\rightarrow $ 6 scenario, demonstrating its consistent superiority in adapting and learning across different scenarios in the PAMAP2 dataset. TrC and DANN demonstrate respectable performances, with DANN leading in the 1 $\rightarrow $ 6 and TrC leading in 5 $\rightarrow $ 6 scenarios, hinting at its capability to adapt well in certain conditions. For instance, TrC and DANN might be better suited to situations where the variance between training and testing data is minimal or where the data patterns are more consistent and less complex. SA and CORAL perform poorly compared to other methods, with accuracy mainly in the range between 30\%-40\%, reflecting their struggle to adapt to this dataset.

\begin{figure}[h!]
\centering
\includegraphics[width=\columnwidth]{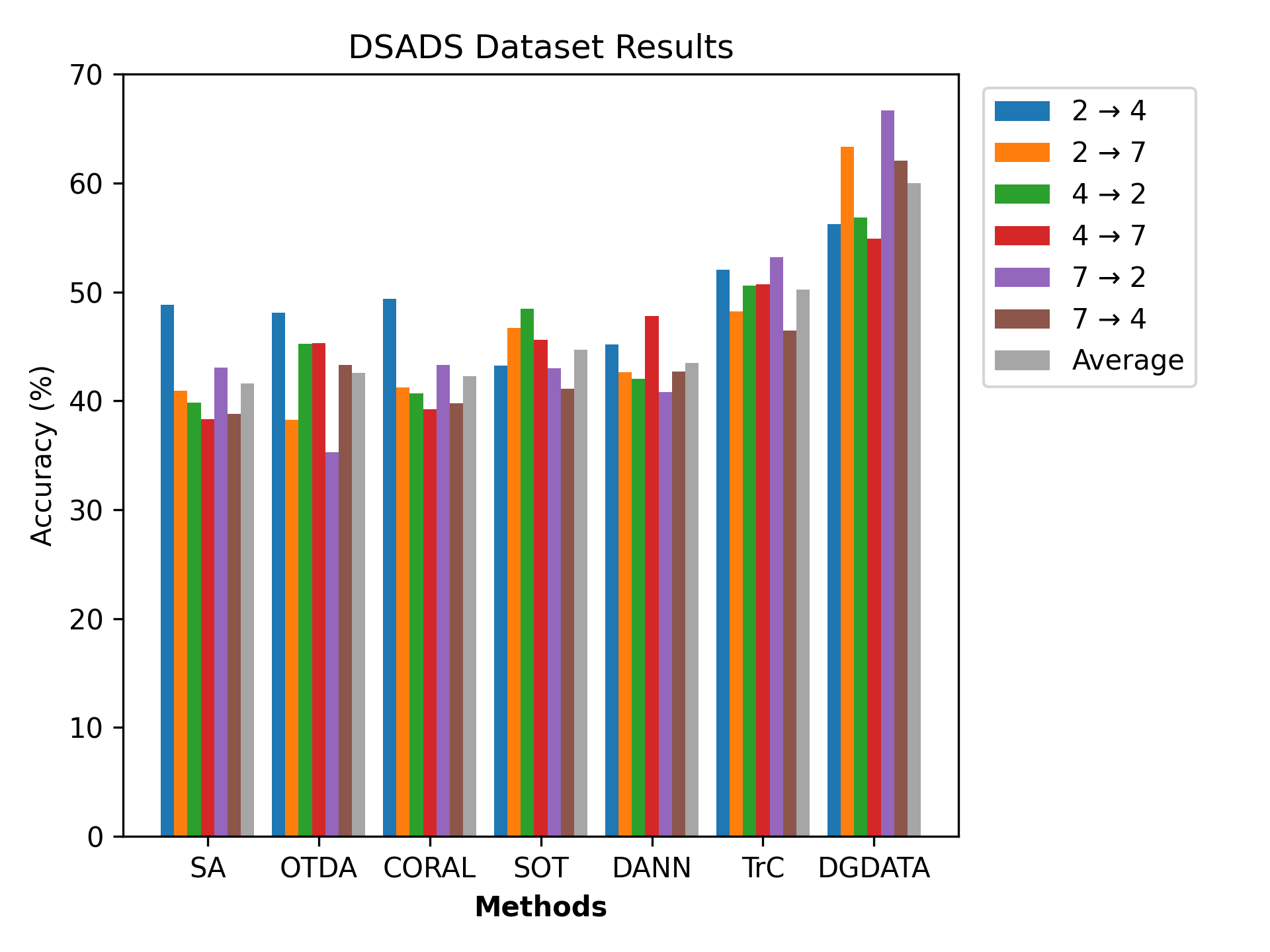}
\caption{DSADS dataset classification results.\label{DSADS_Dataset_Results}}
\end{figure}

On the DSADS dataset (see Figure~\ref{DSADS_Dataset_Results}), DGDATA outperforms all other methods, achieving the highest accuracy in all scenarios. It peaks at 66.69\% in the 7 $\rightarrow $ 2 scenario, demonstrating its robustness and adaptability in the DSADS dataset. TrC also demonstrates a reasonable performance better than the other methods, with accuracy primarily in the range between 40\%-50\%, but still not as efficient as the DGDATA. Other methods show mixed results, with very limited strength in dealing with time series-based cross-user HAR. Their performances are generally middling, with some minor variations.

Across all three datasets, DGDATA consistently achieves the highest scores in all test scenarios. These results demonstrate that our model is highly adaptable, robust, and effective in handling cross-user HAR tasks. TrC consistently performs reasonably well across all datasets, indicating its capability to adapt to different scenarios, though it does not match the performance of DGDATA. SA, OTDA, CORAL, SOT, and DANN show varied and mixed performances across different datasets and scenarios. This suggests that their adaptability and effectiveness are conditional, depending on specific dataset characteristics and scenarios. Several methods, notably SA and CORAL, struggle in certain datasets, suggesting a need for further refinement and enhancement to increase their adaptability and effectiveness in time series data with temporal relation knowledge. The trends and patterns observed are consistent across the different datasets, indicating that the proposed method is reliable in various application scenarios involving different cross-user HAR tasks.

\begin{figure}[h!]
\centering
\includegraphics[width=\columnwidth]{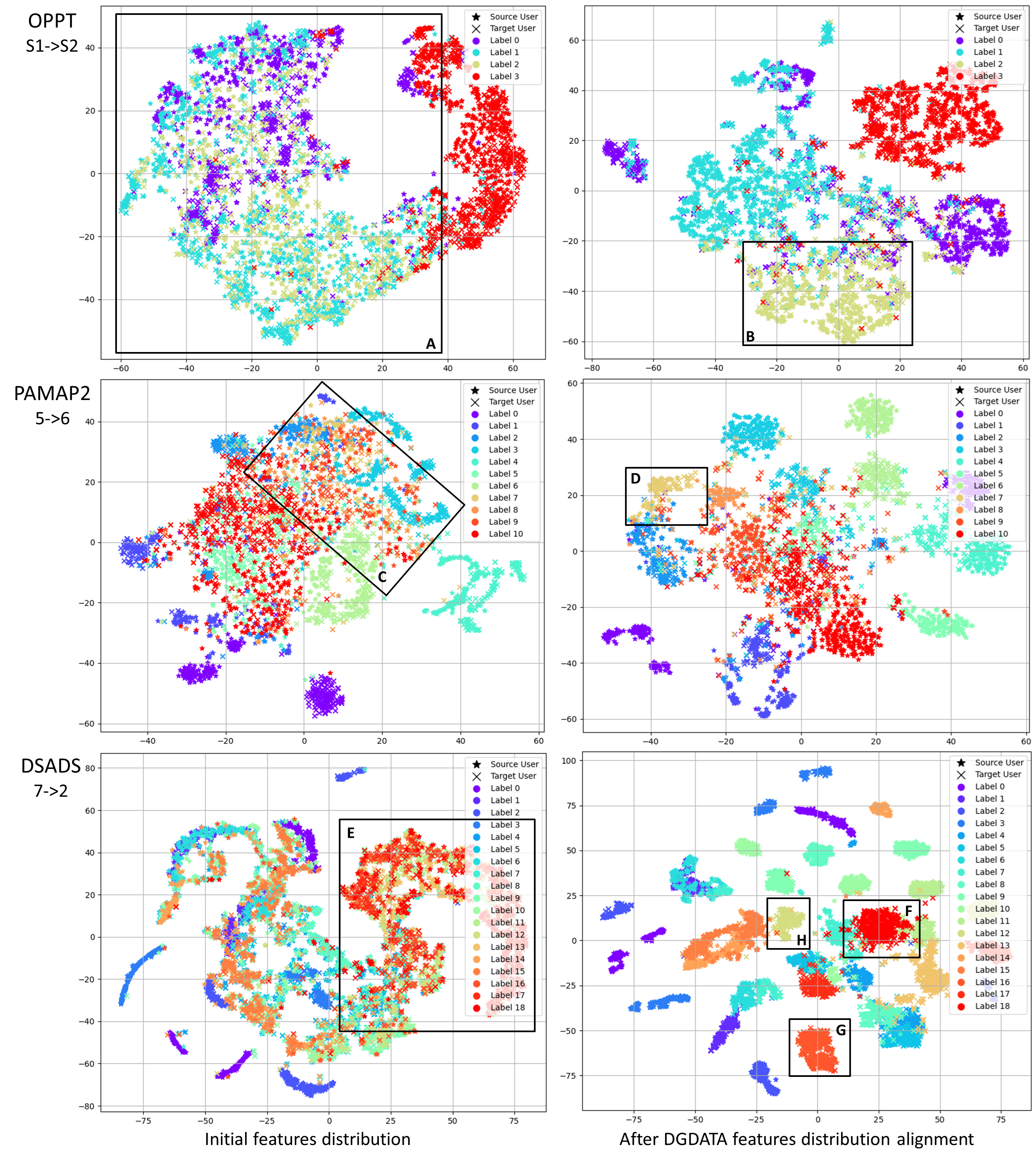}
\caption{t-SNE visualization of features before and after DGDATA adaptation on three datasets.\label{tSNE_all}}
\end{figure}

Figure~\ref{tSNE_all} is the t-SNE embedding of the distributions of the initial features, and the DGDATA learned features after the features distribution mapping for cross-user HAR on three tasks (i.e. OPPT dataset user S1 to S2, PAMAP2 user 5 to 6, and DSADS user 7 to 2 tasks). Different shapes represent different users, and each class of activity is demonstrated by a distinct colour. We visualize the features before and after adaptation to validate whether our proposed DGDATA can effectively acquire features that are not specific to any user for the purpose of classifying activities. 

We notice that in the DGDATA model, features from source and target users blend together as the black square A $\rightarrow $ B, and C $\rightarrow $ D in Figure~\ref{tSNE_all}, showing that these features are not influenced by individual users. Furthermore, after aligning the features using DGDATA, the clusters become clearer and more organized compared to the initial feature distribution as the black square E $\rightarrow $ F, G, H in Figure~\ref{tSNE_all}. This alignment allows samples from the same activity class to cluster into related sub-activities, which are more cohesive and accurately capture common sub-activities across different users, taking into account the temporal relations of these activities. This indicates that DGDATA effectively understands the fundamental structure with temporal relation knowledge. The enhanced clarity in the sub-activity groupings suggests that the model successfully identifies inherent patterns and similarities in the temporal-related sub-activities, regardless of user variability. This finding is crucial as it shows the model's ability to generalize across different users by making use of temporal relation knowledge, making it robust for activity classification.

\subsection{Effect of temporal relation knowledge}

Here, we analyze the effect of temporal relation knowledge. The studies under discussion utilize the OPPT, PAMAP2, and DSADS datasets. The necessity of the usage of temporal relation knowledge is explored. Confusion matrix analysis is employed to compare SOT, DANN, and our DGDATA method for analyzing where each method excels and where it might be falling short. SOT represents a conventional domain adaptation technique of high caliber, while DANN is noted for its deep domain adaptation approach. It is important to note, however, that neither SOT nor DANN were specifically engineered for adapting to time series domains, a gap that our DSADS method seeks to address.

\begin{figure}[h!]
\centering
\includegraphics[width=\columnwidth]{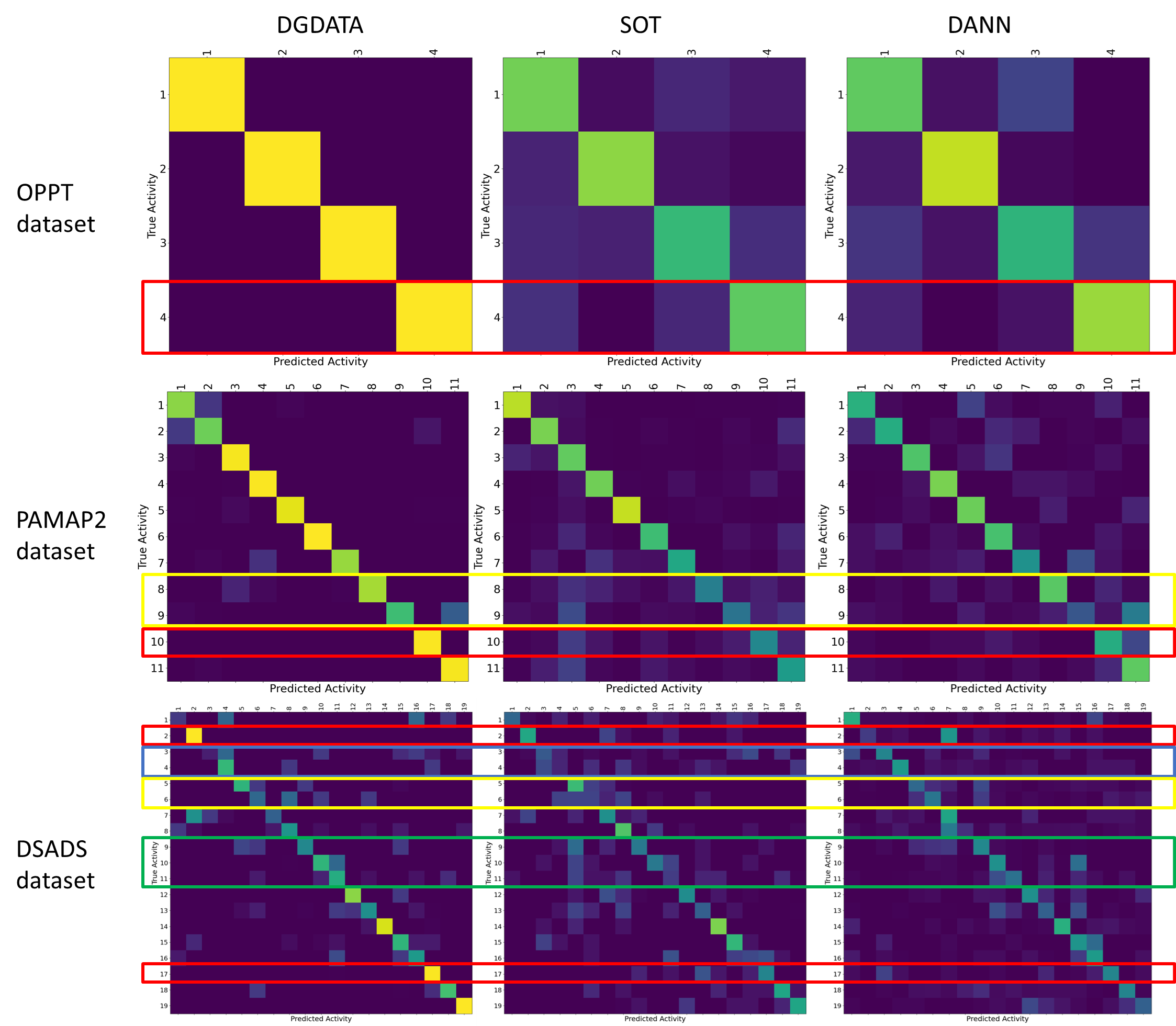}
\caption{Confusion matrices of DGDATA, SOT and DANN methods for the average performance of tasks in the DSADS, PAMAP2 and OPPT datasets.\label{cm_results}}
\end{figure}

In Figure~\ref{cm_results}, we examine the confusion matrices for the DGDATA, SOT, and DANN methods, assessing their collective performance on the DSADS, PAMAP2, and OPPT datasets. These matrices are arranged such that both axes represent indices of various activity classes, with specific index-to-activity correlations provided in Table~\ref{tab_datasets_info}. The colour gradient within these matrices is key; a shift towards warmer tones, particularly yellow, indicates higher accuracy. On the matrices, the x-axis corresponds to the activities as predicted by the methods, and the y-axis represents the actual true activities. A critical feature to observe is the diagonal line from the top left to the bottom right, which symbolizes correct predictions or true positives. This is where the predicted activities align with the true activities, serving as an indicator of the accuracy and effectiveness of each method in classifying the activities.

All methods, notably demonstrated in these analyses, exhibit pronounced values along their diagonals. This signifies a predominant accuracy in activity recognition across each method, underscoring their efficiency in cross-user scenarios. Common activities such as 'lying', 'sitting', 'standing', 'running', and 'cycling' are adeptly identified by all methods. However, DGDATA distinguishes itself by accurately identifying complex activities like 'ironing', 'vacuum cleaning', and 'Nordic walking' with a superior precision compared to its competitors. Additionally, while the SOT and DANN methods show a more dispersed classification spectrum, DGDATA exhibits a more condensed and clearer classification, as the red squares shown in Figure~\ref{cm_results}, attributable to its handling of temporal states and sub-activity extraction.

Activities with distinct features, such as "exercising on cross trainer", are well recognized by all methods. Moreover, activities like 'lying', 'walking', and 'running' demonstrate high accuracy across all methods, likely due to their unique movement patterns. For instance, 'lying' typically involves minimal motion, whereas 'walking' and 'running' are differentiated by pace and stride, with walking exhibiting a rhythmic and slower movement compared to the faster-paced running. However, all methods exhibit confusion when distinguishing between similar activities. For example, different types of walking ('walking in parking lot', 'walking on treadmill flat', and 'walking on treadmill inclined') are often conflated as the green square in Figure~\ref{cm_results}. Misclassifications also occur in activities with similar temporal patterns or stationary postures, such as 'ascending stairs' versus 'descending stairs' as the yellow squares in Figure~\ref{cm_results}, or 'lying on the back' versus 'lying on the right' as the blue square in Figure~\ref{cm_results}.

Interestingly, DGDATA also outperforms SOT and DANN in identifying low-motion activities like 'standing', 'sitting', and 'lying'. This can be attributed to DGDATA’s nuanced capture of temporal relation context that mitigated the effect of subtle sensor data variation. Furthermore, DGDATA proves superior in distinguishing dynamic activities and those with complex temporal relations, such as 'playing basketball', 'jumping', and 'rowing', thanks to its ability to capture diverse motion patterns and rhythmic sequences embedded in the corresponding time series data.

In certain activities, people often perform transitional movements that, while not the main focus of the activity, occur regularly enough to be significant. For example, during ironing, individuals frequently adjust the garment they are ironing. Similarly, when vacuum cleaning, they may need to reposition the vacuum cleaner occasionally. These secondary movements can sometimes resemble other activities, causing potential confusion. Unlike SOT and DANN, DTSDA effectively captures the knowledge of these temporal relations, which helps in better understanding and identifying these transitional movements, which leads to a better classification result for the actual activity.

The integration of temporal relation knowledge in DGDATA helps to minimize the differences in activity patterns between different users, particularly in dynamic activities that have complex temporal relations. By extracting and understanding the time-related information within sequential data, the process of recognizing activities becomes more robust and accurate. This approach is especially effective for activities that have rhythmic or sequential traits. The use of temporal knowledge not only improves the ability to recognize various activities but also mitigates the subtle nuances in different human behaviours.

\section{Conclusion}

This research introduces the DGDATA method, an innovative methodology designed for time series-based cross-user HAR via domain adaptation. Uniquely, our method departs from conventional approaches by leveraging temporal relations embedded in time series, which offers better ability to extract user-invariant temporal features rather than following the common assumption of $\displaystyle i.i.d.$ samples. Utilizing a deep generative model architecture, augmented with a Temporal Attention mechanism and adversarial learning strategies, DGDATA achieves effective cross-user HAR. Importantly, this method facilitates the extraction of user-invariant temporal relations, a key factor contributing to its performance.

The effectiveness of the DGDATA method was evaluated through experiments conducted on three public HAR datasets. Our method surpassed some other models in the context of cross-user HAR, thereby underscoring the practical utility promise of our approach. These encouraging outcomes establish DGDATA as a solid step to solving time series-specified domain adaptations such as cross-user HAR. Specifically, incorporating temporal relation into domain adaptation offers a reasonable pathway to reduce the data distribution discrepancies between source and target users in HAR applications.

In future work, we plan to test DGDATA on more complex or less structured activities, which could provide insights into its adaptability and robustness. Activities with higher complexity in their temporal relation patterns would be a good challenge for the method. Moreover, extending experiments to include a wider range of user groups, particularly those with varying levels of physical ability or different demographic characteristics, could help in understanding how well DGDATA generalizes across diverse populations.

 \bibliographystyle{elsarticle-num} 
 \bibliography{ref}

\section*{Biography}

\noindent
\textbf{Xiaozhou Ye} received the M.Sc. degree from Hohai University, Nanjing, China, in 2017. He is currently pursuing the Ph.D. degree with the Department of Electrical, Computer, and Software Engineering, The University of Auckland, Auckland, New Zealand. He worked in industries for business intelligence and data analysis in New Zealand from 2018 to 2020. His current research interests include transfer learning, human activity recognition (HAR), and pervasive healthcare systems.

\begin{figure}[h]
\centering
\includegraphics[width=1in,height=1.25in,keepaspectratio]{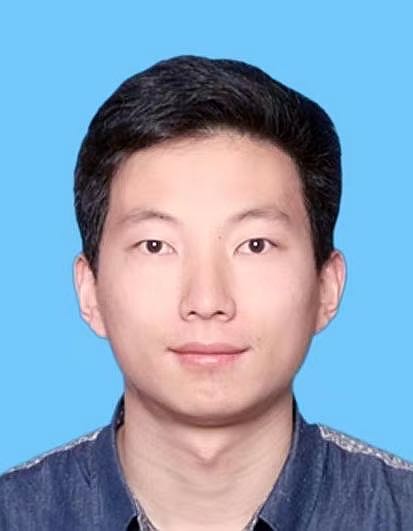}
\end{figure}

\noindent
\textbf{Kevin I-Kai Wang} (Member, IEEE) received the B.E. degree (Hons.) in computer systems engineering and the Ph.D. degree in electrical and electronics engineering from the Department of Electrical and Computer Engineering, The University of Auckland, Auckland, New Zealand, in 2004 and 2009, respectively. He was a Research Engineer designing commercial home automation systems and traffic sensing systems from 2009 to 2011. He is currently a Senior Lecturer with the Department of Electrical and Computer Engineering at the University of Auckland. His current research interests include wireless sensor network-based ambient intelligence, pervasive healthcare systems, human activity recognition (HAR), behaviour data analytics, and biocybernetic systems.

\begin{figure}[h]
\centering
\includegraphics[width=1in,height=1.25in,keepaspectratio]{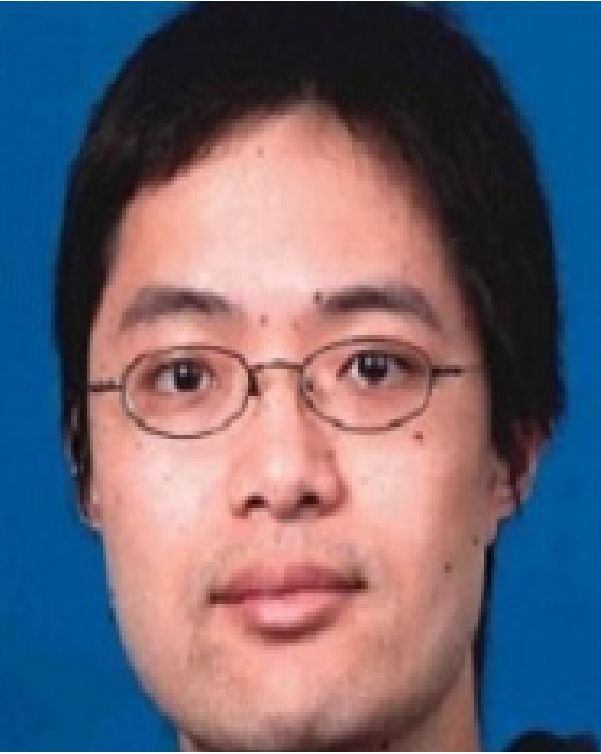}
\end{figure}

\end{document}